\documentclass[a4paper,fleqn]{cas-dc}

\usepackage[numbers]{natbib}

\def\tsc#1{\csdef{#1}{\textsc{\lowercase{#1}}\xspace}}
\tsc{WGM}
\tsc{QE}

\begin{document}
\let\WriteBookmarks\relax
\def\floatpagepagefraction{1}
\def\textpagefraction{.001}

\shorttitle{}    

\shortauthors{}  

\title [mode = title]{Vision-Language Model Guided Image Restoration}  



%

\author[1]{Cuixin Yang}
\ead{cuixin.yang@connect.polyu.hk}

\credit{Writing - original draft, Conceptualization, Methodology, Software, Visualization, Validation, Data curation}

\author[1]{Rongkang Dong}
\ead{rongkang97.dong@connect.polyu.hk}
\credit{Writing - methodology, Validation}

\author[1]{Kin-Man Lam}
\ead{enkmlam@polyu.edu.hk}
\credit{Writing - review \& editing, Supervision}
\cormark[1]

\affiliation[1]{organization={Department of Electrical and Electronic Engineering},
                addressline={The Hong Kong Polytechnic University}, 
                city={Hong Kong},
                country={China}}

\cortext[1]{Corresponding author at: Department of Electrical and Electronic Engineering, The Hong Kong Polytechnic University, Hong Kong, China.}



\begin{abstract}
Many image restoration (IR) tasks require both pixel-level fidelity and high-level semantic understanding to recover realistic photos with fine-grained details. However, previous approaches often struggle to effectively leverage both the visual and linguistic knowledge. Recent efforts have attempted to incorporate Vision-language models (VLMs), which excel at aligning visual and textual features, into universal IR. Nevertheless, these methods fail to utilize the linguistic priors to ensure semantic coherence during the restoration process. To address this issue, in this paper, we propose the Vision-Language Model Guided Image Restoration (VLMIR) framework, which leverages the rich vision-language priors of VLMs, such as CLIP, to enhance IR performance through improved visual perception and semantic understanding. Our approach consists of two stages: VLM-based feature extraction and diffusion-based image restoration. In the first stage, we extract complementary visual and linguistic representations of input images by condensing the visual perception and high-level semantic priors through VLMs. Specifically, we align the embeddings of captions from low-quality and high-quality images using a cosine similarity loss with LoRA fine-tuning, and employ a degradation predictor to decompose degradation and clean image content embeddings. These complementary visual and textual embeddings are then integrated into a diffusion-based model via cross-attention mechanisms for enhanced restoration. Extensive experiments and ablation studies demonstrate that VLMIR achieves superior performance across both universal and degradation-specific IR tasks, underscoring the critical role of integrated visual and linguistic knowledge from VLMs in advancing image restoration capabilities.
\end{abstract}


\maketitle

\section{Introduction}
Image restoration (IR) is a fundamental and popular computer vision task, aimed at recovering high-quality (HQ) images from their low-quality (LQ) counterparts \cite{guo2025one, yang2025geometric}. It has a wide range of applications spanning surveillance, autonomous driving, and multimedia entertainment. IR tasks, such as denoising \cite{tirer2018image}, super-resolution \cite{yang2023improving}, and inpainting \cite{quan2024deep}, seek to recover fine-grained details while preserving semantic integrity. However, most existing IR methods \cite{zhang2018ffdnet, kwon2022learning}, which rely on exploiting low-level image statistics with diverse network architectures \cite{he2016identity, vaswani2017attention} and training losses \cite{goodfellow2020generative}, struggle to balance pixel-level fidelity with high-level semantic coherence. 

The emergence of vision-language models (VLMs) \cite{zhang2024vision}, which integrate visual and linguistic understanding, offers a promising opportunity to address the above challenges by leveraging multimodal knowledge to guide restoration processes. VLMs, such as CLIP \cite{potlapalli2023promptir} and DALL-E \cite{ramesh2021zero}, pretrained on large-scale datasets of image-text pairs, enable the alignment of visual features with corresponding textual descriptions in a shared embedding space. With their advanced capabilities in both visual and linguistic understanding, VLMs have been successfully applied to various tasks \cite{abdelfattah2023cdul, qian2024online}, including image-text retrieval, object detection, and segmentation. However, applying VLMs to IR tasks presents several challenges. One major issue is domain mismatch: VLMs are typically pretrained on clean, HQ images paired with descriptive captions, making them less effective when dealing with degraded, LQ images. Recent advancements \cite{luo2023controlling, luo2024photo, wei2025leveraging} have demonstrated that VLMs can be incorporated into IR to further enhance the performance. These methods generally involve training an image controller to predict the degradation type of the LQ image, thereby enabling the separation of clean image content and degradation embeddings, which are then integrated into the IR model. However, these methods often overlook the use of linguistic priors, such as textual embeddings derived from image captions, which can provide high-level semantic understanding and coherence during the IR process. Another approach, known as Agentic IR \cite{chen2024restoreagent, zhu2024intelligent, zuo20254kagent}, utilizes VLMs as agents to provide descriptions about image degradation and schedule a restoration plan, leveraging their advanced reasoning and planning capabilities. However, Agentic IR primarily focuses on generating descriptions or instructions, instead of exploiting the fine-grained feature-level priors encoded in VLM embeddings. This limits the full potential of vision-language knowledge from VLMs to be utilized in IR modeling.

To fully exploit the vision-language knowledge embedded in VLMs, in this paper, we introduce the Vision-Language Model Guided Image Restoration (VLMIR) framework. Inspired by DA-CLIP \cite{luo2023controlling}, and using CLIP \cite{radford2021learning}, our method learns both image content embeddings and image caption embeddings, which form complementary visual-linguistic information, as well as degradation embeddings. These are then employed to guide the downstream image restoration process. Specifically, VLMIR consists of two main stages: VLM-based feature extraction and diffusion-based image restoration. 

In the first stage, we utilize the bootstrapped vision-language framework BLIP \cite{li2022blip} to generate captions for LQ images (referred to as LQ captions) and their corresponding HQ images (GT captions). GT captions, generated from clean HQ images, are assumed to be accurate descriptions. Due to degradation, LQ captions may be misleading and not completely accurate, as shown in Fig.~\ref{overview}. Thus, we align the LQ caption embedding and the GT caption embedding using cosine similarity loss, fine-tuning the LQ caption text encoder with LoRA \cite{hu2022lora}. Additionally, a degradation predictor is trained using contrastive learning to identify the degradation type, enabling the image encoder to generate clean image embeddings \cite{luo2023controlling}. 

In the second stage, the extracted complementary visual and textual features are integrated into a diffusion-based image restoration model through cross-attention mechanisms. To ensure stable training, we adopt IR-SDE \cite{luo2023image} as the backbone model.

In summary, the main contributions of this paper are as follows:
\begin{itemize}
    \item We propose the Vision-Language Model Guided Image Restoration (VLMIR) framework, which utilizes complementary visual and linguistic priors from VLMs to guide the image restoration process.
    \item We introduce a method to learn both image content and image caption embeddings based on CLIP, aligning LQ captions and GT captions via cosine similarity loss with LoRA fine-tuning, and employing contrastive learning to obtain clean image content and degradation embeddings.
    \item Extensive experiments and ablation studies validate the effectiveness of the proposed method, demonstrating superior performance in both universal image restoration and degradation-specific image restoration tasks, based on both quantitative and qualitative evaluations. 
\end{itemize}

\section{Related Works}
\subsection{Image Restoration}
Image restoration (IR) is a classical and pivotal low-level computer vision task. Early IR methods primarily relied on handcrafted priors \cite{weiss2007makes} and mathematical models \cite{rudin1992nonlinear} to address image degradation. Techniques such as median filtering \cite{justusson2006median} and Wiener filtering \cite{vaseghi1996wiener} for denoising assume specific noise distributions but often result in oversmoothing and loss of detail. Methods like total variation regularization \cite{rudin1992nonlinear} and sparse coding \cite{elad2006image} impose constraints, e.g., smoothness and sparsity, to reconstruct images, but they struggle with complex degradations. 

The advent of deep learning has revolutionized IR by introducing data-driven models that learn complex mappings from degraded LQ images to clean HQ images. Models such as DnCNN \cite{zhang2017beyond} for denoising and SRResNet \cite{ledig2017photo} for super-resolution use deep convolutional architectures to learn hierarchical features, enabling the restoration of fine details with high fidelity and significantly improving overall image quality. More recently, Transformer-based architectures, such as SwinIR \cite{liang2021swinir}, have demonstrated enhanced performance by capturing global representations through long-range dependency modeling. 

Emerging generative models, including generative adversarial networks (GANs) \cite{goodfellow2020generative} and diffusion models \cite{rombach2022high}, further enhance perceptual quality by generating realistic textures and structures, particularly in tasks like inpainting and super-resolution. Although these methods excel in specific degradation scenarios, thanks to optimized training, tailored architectures with loss functions, they often struggle to generalize across diverse IR tasks within a single model due to out-of-distribution degradations \cite{li2022all}. The need for separate training pipelines introduces inefficiencies when addressing complex and varied real-world applications.

\subsection{Universal Image Restoration}
Recently, universal IR models, also known as all-in-one IR models, have been introduced to address the limitations of task-specific approaches \cite{li2022all, park2023all, potlapalli2023promptir, jiang2024autodir}. These models \cite{valanarasu2022transweather,zhang2025perceive,zeng2025all} aim to handle multiple isolated or mixed degradation types within a unified framework \cite{jiang2024survey}, often explicitly or implicitly classifying degradation types \cite{luo2023controlling, hu2025universal}. 
Early universal IR methods tackled diverse degradation challenges using techniques such as decoupled learning \cite{fan2019general}, tailored encoders \cite{li2020all}, or specialized decoder heads \cite{chen2021pre}. More recent advancements have shifted toward more integrated and robust solutions.
For example, AirNet \cite{li2022all} proposes contrastive learning of degradation representations using positive and negative samples, which are then used to guide the restoration process. PromptIR \cite{potlapalli2023promptir} introduces a plug-and-play prompt module that implicitly predicts the degradation type and injects the encoded degradation features into the restoration network. InstructIPT \cite{tian2024instruct} presents a simple yet effective weight modulation method, arguing that weight modulation is more suitable than feature modulation for highly unrelated specific tasks. 
More recently, external large-scale models \cite{radford2021learning} and generative priors \cite{ho2020denoising} have been introduced to further enhance performance \cite{luo2023controlling, zheng2024selective, luo2024photo, wei2025leveraging}. Vision-language models (VLMs), such as CLIP \cite{radford2021learning}, training on web-scale datasets, excel in embedding aligned visual and linguistic features.
InstructIR \cite{conde2024instructir} utilizes GPT-4 to generate user prompts and adopts these instructions as a control mechanism to recover HQ images. 
DA-CLIP \cite{luo2023controlling} employs CLIP \cite{radford2021learning} to learn representations of both degradation and image content, which are then utilized to facilitate the restoration of LQ images. 
To broaden the application scope of DA-CLIP, the authors leverage a synthetic degradation pipeline to simulate wild degradations, enabling its application to wild IR scenarios \cite{luo2024photo}. 
Building on DA-CLIP, the authors of \cite{wei2025leveraging} propose a real-world adverse weather removal approach using pseudo-ground truth generation, and utilizes the RainMix \cite{guo2021efficientderain, ba2022not} data augmentation method to merge synthetic rain streak masks into degraded images.

However, these methods remain limited in their use of the unified visual and linguistic representations offered by powerful VLMs. They typically integrate only the visual and degradation embeddings into the IR model, while ignoring the complementary linguistic embeddings derived from image captions.
In contrast, in this paper, we propose a VLM-guided framework for IR that jointly addresses both visual perception and linguistic coherence throughout the restoration process.

\subsection{Vison-Language Models}
Vision-language models (VLMs) combine computer vision and natural language processing to process and understand multimodal data, including images, videos, and texts. The core of VLMs is to align representations of visual perception and linguistic representations within a shared embedding space. By learning from visual and textual contexts simultaneously, VLMs are endowed with excellent zero-shot capabilities and support a wide array of applications, such as image captioning, visual question answering, and multimodal reasoning. CLIP learns to align image and text embeddings by contrastive learning. BLIP \cite{li2022blip} utilizes noisy web data by bootstrapping the synthetic captions. Flamingo \cite{alayrac2022flamingo} is trained on large-scale image-text datasets containing arbitrarily interleaved text and images. Based on LLaMA \cite{touvron2023llama}, LLaVA \cite{liu2023visual} integrates a pretrained language model with a vision encoder to process, understand and generate multimodal outputs with relatively lower computational demands. DALL-E \cite{ramesh2021zero} and Stable Diffusion \cite{rombach2022high} are based on generative models, such as autoregressive frameworks \cite{gregor2014deep} and diffusion models \cite{ho2020denoising} to generate images from textual prompts. 

Recently, VLMs have been introduced into the field of universal IR. Methods like \cite{luo2023controlling, luo2024photo, wei2025leveraging} utilize CLIP \cite{radford2021learning} to obtain clean image content and degradation embeddings as prompts to assist the IR process. On the other hand, Agentic IR methods utilize LLMs/VLMs to schedule restoration plans rather than exploiting the feature-level priors. In this paper, we will effectively exploit the feature-level visual and linguistic priors of VLMs to guide the IR process.

\section{Methodology}
\subsection{Overview}
\begin{figure*}[!t]
\centering
\includegraphics[width=\textwidth]{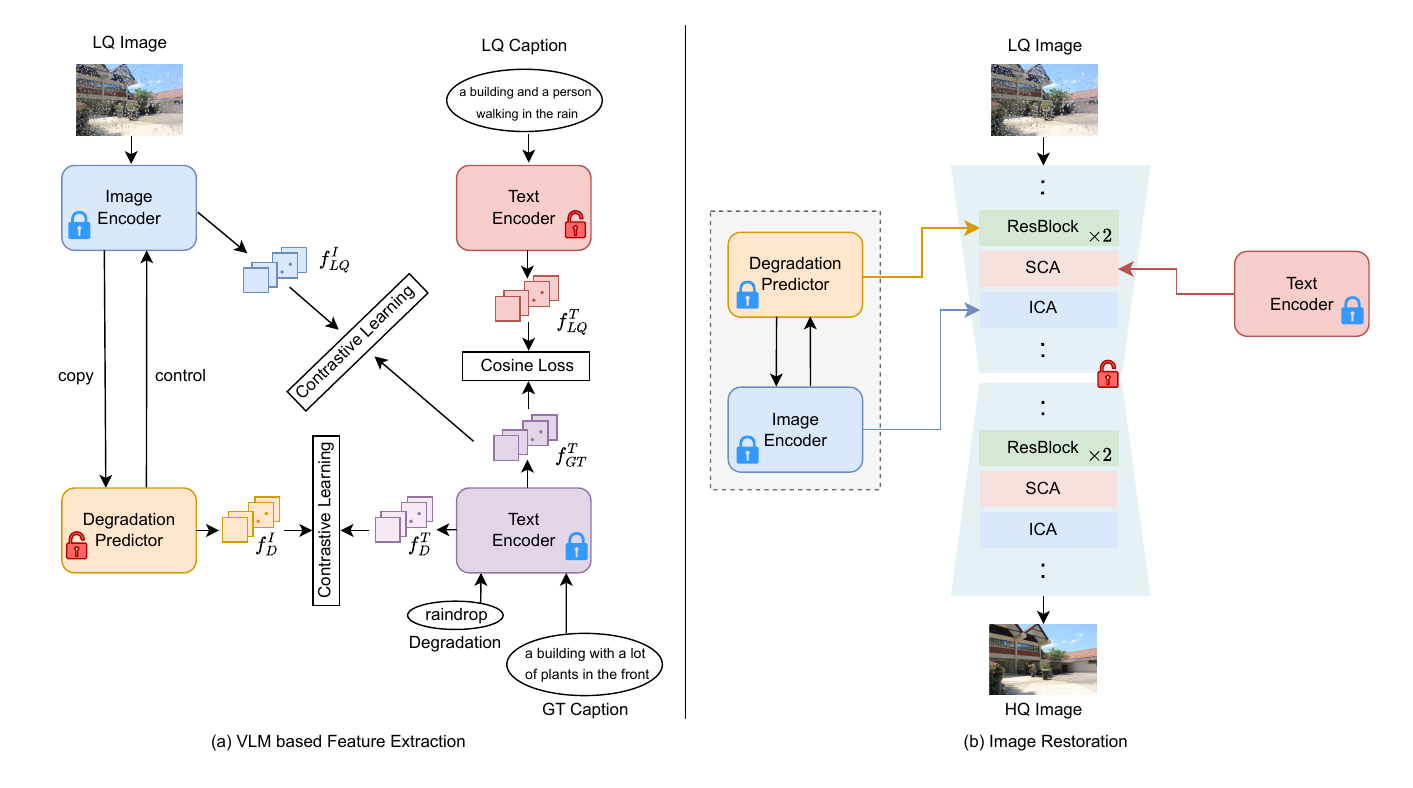}
\caption{Overview of the proposed Vision-Language Model Guided Image Restoration (VLMIR) framework.}
\label{overview}
\end{figure*}
An overview of the proposed method is illustrated in Fig.~\ref{overview}. The training process is divided into two stages: VLM-based feature extraction and image restoration. 
In the first stage, VLM-based feature extraction aims to obtain a clean image embedding and accurately predict the degradation type of the LQ image. As shown in Fig.~\ref{overview}, this is achieved by aligning the embedding of the LQ image $f^{I}_{LQ}$ with that of the GT caption $f^{T}_{GT}$, and by matching the predicted degradation $f^{I}_{D}$ with the degradation embedding generated by the text encoder $f^{T}_{D}$. To further enhance the model's semantic understanding, we align the text embedding of the LQ caption $f^{T}_{LQ}$ with that of the GT caption $F^{T}_{GT}$. Note that both synthetic GT and LQ captions are generated using the bootstrapped vision-language framework BLIP \cite{li2022blip}. These features are then incorporated into the training of the downstream image restoration model to enhance its performance.

\subsection{VLM-based Feature Extraction}
Linguistic information has proven effective in many computer vision tasks \cite{liu2023polyformer, yang2023reco}, though its application in image restoration remains underexplored. To enhance contextual understanding and generate higher-quality images, we propose incorporating both visual and linguistic representations during the restoration of degraded images. We adopt the pretrained CLIP \cite{potlapalli2023promptir} model as the backbone for feature extraction. CLIP is a multimodal vision-language model that learns both images and text and bridges them through contrastive learning. The latent features extracted from CLIP's image and text encoders are designed to provide well-aligned representations of image-text pairs. As shown in Fig.~\ref{overview}, the textual description of degradation and the GT Caption are input into the text encoder. The degradation predictor is trained to classify the degradation type of the LQ image through contrastive learning between the predicted degradation $f^{I}_{D}$ and the degradation embedding generated by the text encoder $f^{T}_{D}$. This enables the image encoder to generate a clean image embedding $f^{I}_{LQ}$ guided by the GT caption embedding $f^{T}_{GT}$. The contrastive loss is formulated as follows:
\begin{equation}
    L_{c}(m, n) = -\frac{1}{K}\sum_{i}\text{log}(\frac{e^{m^{T}_{i}n_{i}/s}}{\sum_{j}e^{m^{T}_{i}n_{j}/s}}),
\end{equation}
where $m$ and $n$ denote batches of normalized vectors of image-text embedding pairs, i.e., $f^{I}_{LQ}$ \& $f^{T}_{GT}$ and $f^{I}_{D}$ \& $f^{T}_{D}$, such that $m_{i}$ and $n_{i}$ are embeddings of a positive pair, while $n_{j}$ represents embeddings of all other samples (including negatives); $K$ is the number of paired features in the training batches; and $s$ is a learnable factor that adjusts the contrastive strength.

Semantic descriptions have shown their effectiveness in various computer vision tasks \cite{liu2023polyformer, yang2023reco}. To enrich the textual information available to the downstream image restoration models, we introduce an additional text encoder to encode the caption of the LQ image (LQ Caption). Due to the low visual quality and distortion in degraded images, the generated LQ captions may not precisely describe the image content, introducing noise (see in Fig.~\ref{overview}). To correct the latent embedding of the LQ captions, i.e., $f^{T}_{LQ}$, we align them with the latent embedding of the corresponding GT captions, i.e., $f^{T}_{GT}$. Specifically, the initial parameters of the LQ caption text encoder are copied from the pretrained CLIP text encoder. We then fine-tune it using LoRA with the cosine similarity loss between $f^{T}_{LQ}$ and $f^{T}_{GT}$ defined as follows:
\begin{equation}
    L_{cos}(m, n) = 1 - \text{cos}(m, n),
\end{equation}
where $\text{cos}(\cdot, \cdot)$ denotes the cosine similarity between two normalized vectors $m$ and $n$. To summarize, the total training loss for the first stage is expressed as follows:
\begin{equation}
L_{1} = L_c{(f^{I}_{LQ}, f^{T}_{GT})} + L_{c}{(f^{I}_{D}, f^{T}_{D})} + \alpha\times L_{cos}(f^{T}_{LQ}, f^{T}_{GT}),
\end{equation}
where $\alpha$ is a weighting factor for the cosine similarity loss.
This framework fully leverages both visual and textual features, which can be integrated into the downstream image restoration model to enhance its performance.

\subsection{Image Restoration}
In the image restoration stage, we adopt IR-SDE \cite{luo2023image}, which is a U-net-based diffusion model, as our backbone. All parameters of the backbone model are trained from scratch, while the image encoder, text encoder, and degradation predictor remain fixed.
To utilize both the text embeddings of LQ captions and the clean image embeddings, we introduce two mechanisms into the diffusion model: semantic cross-attention (SCA) and image cross-attention (ICA). The ICA modules are placed after the SCA modules. By conditioning the diffusion model on both text and image embeddings, it gains enhanced semantic understanding of the image, thereby improving restoration performance. In addition, the degradation embedding generated by the degradation predictor is used to classify the specific degradation type of the degraded LQ image. This allows the model to handle various types of degradation using a single model. Specifically, the degradation embedding is injected into the residual blocks and fused using a learnable prompt to generate scale and shift factors that transform the input features \cite{luo2023controlling}. Given the noisy state $x_{t}$ at time $t$, and the LQ image $\mu$ as the condition, the diffusion model is denoted as $\epsilon_{\theta}(x_{t}, \mu, t, f^{T}_{LQ}, f^{I}_{LQ}, f^{T}_{D})$. The diffusion model is trained from scratch using likelihood learning, following the stable training strategy of IR-SDE \cite{luo2023image}.

\begin{table*}
\centering
\caption{Quantitative comparison of different universal image restoration methods on various image restoration tasks. Best results are highlighted in red, and second-place results are marked in blue.}
\begin{tabular}{c|c|cccccc}
\hline
Method   & Degradation                & PSNR                         & SSIM                          & Y-PSNR                       & Y-SSIM                        & LPIPS                         & FID                          \\ \hline\hline
AirNet \cite{li2022all}   &                            & 22.09                        & 0.8218                        & 23.66                        & 0.8732                        & 0.0757                        & 119.50                       \\
PromptIR \cite{potlapalli2023promptir} &                            & 28.01                        & 0.8771                        & 29.83                        & 0.9134                        & {\color[HTML]{FE0000} 0.0429} & 72.33                        \\
IR-SDE \cite{luo2023image}   &                            & 27.89                        & 0.8707                        & 29.57                        & 0.9039                        & 0.0633                        & 32.37                        \\
DA-CLIP \cite{luo2023controlling}  &                            & {\color[HTML]{3166FF} 29.09} & {\color[HTML]{3166FF} 0.8856} & {\color[HTML]{3166FF} 30.89} & {\color[HTML]{3166FF} 0.9161} & 0.0568                        & {\color[HTML]{FE0000} 21.72} \\
VLMIR    & \multirow{-5}{*}{Raindrop} & {\color[HTML]{FE0000} 29.36} & {\color[HTML]{FE0000} 0.8866} & {\color[HTML]{FE0000} 31.14} & {\color[HTML]{FE0000} 0.9175} & {\color[HTML]{3166FF} 0.0553} & {\color[HTML]{3166FF} 25.99} \\ \hline
AirNet \cite{li2022all}   &                            & 27.45                        & 0.9475                        & 28.96                        & 0.9697                        & {\color[HTML]{FE0000} 0.0150} & 29.59                        \\
PromptIR \cite{potlapalli2023promptir} &                            & 27.58                        & 0.9479                        & 29.13                        & 0.9699                        & {\color[HTML]{3166FF} 0.0162} & 26.25                        \\
IR-SDE \cite{luo2023image}   &                            & 18.67                        & 0.8411                        & 20.11                        & 0.8933                        & 0.0928                        & 19.80                        \\
DA-CLIP \cite{luo2023controlling}  &                            & {\color[HTML]{3166FF} 30.01} & {\color[HTML]{3166FF} 0.9488} & {\color[HTML]{3166FF} 31.62} & {\color[HTML]{3166FF} 0.9742} & {\color[HTML]{343434} 0.0199} & {\color[HTML]{FE0000} 7.45}  \\
VLMIR    & \multirow{-5}{*}{Dehazing}     & {\color[HTML]{FE0000} 30.22} & {\color[HTML]{FE0000} 0.9523} & {\color[HTML]{FE0000} 31.85} & {\color[HTML]{FE0000} 0.9745} & {\color[HTML]{343434} 0.0199} & {\color[HTML]{3166FF} 8.99}  \\ \hline
AirNet \cite{li2022all}   &                            & {\color[HTML]{FE0000} 28.00} & {\color[HTML]{FE0000} 0.7975} & {\color[HTML]{FE0000} 29.67} & {\color[HTML]{FE0000} 0.8214} & {\color[HTML]{FE0000} 0.0747} & 86.85                        \\
PromptIR \cite{potlapalli2023promptir} &                            & 27.06                        & {\color[HTML]{3166FF} 0.7517} & {\color[HTML]{3166FF} 28.88} & {\color[HTML]{3166FF} 0.7884} & {\color[HTML]{3166FF} 0.0965} & 89.19                        \\
IR-SDE \cite{luo2023image}   &                            & 25.54                        & 0.7057                        & 27.18                        & 0.7427                        & 0.1833                        & 77.03                        \\
DA-CLIP \cite{luo2023controlling}  &                            & 25.71                        & 0.7157                        & 27.28                        & 0.7493                        & 0.1766                        & {\color[HTML]{3166FF} 74.09}                        \\
VLMIR    & \multirow{-5}{*}{Denoising}    & {\color[HTML]{3166FF} 25.74} & 0.7171                        & 27.32                        & 0.7507                        & 0.1793                        & {\color[HTML]{FE0000} 71.82} \\ \hline
\end{tabular}
\label{quan1}
\end{table*}

\section{Experiments}
\subsection{Experimental Settings}
\subsubsection{Datasets}
We conduct experiments under two types of image restoration training settings: universal image restoration and degradation-specific image restoration. Degradation-specific image restoration models are trained on datasets containing images with only one specific type of degradation. 
Universal image restoration models, also known as all-in-one image restoration models, are trained on mixed datasets containing multiple types of degradation, aiming to tackle various image restoration tasks within a unified framework. In this work, we evaluate the performance of our model on three image restoration tasks: raindrop removal, dehazing, and denoising. For raindrop removal, we adopt the RainDrop dataset \cite{qian2018attentive}, which contains 861 training images and 58 testing images. For dehazing, the RESIDE-6k dataset \cite{qin2020ffa} is employed. Due to its large size, which may lead to data imbalance, we select the first 1,000 training images and the first 50 testing images to form the training and test datasets. For denoising, we adopted the widely used DIV2K dataset \cite{agustsson2017ntire}, which contains 900 images, as the training dataset, and CBSD68 \cite{martin2001database}, which contains 68 images, as the test dataset. LQ noisy images are synthesized by adding Gaussian noise with a noise level of 50. GT captions and LQ captions are generated using the vision-language model BLIP \cite{li2022blip}. Given the high quality of the GT images, their GT captions are assumed to be accurate and are used as supervision during the training of the first-stage model. 

\subsubsection{Evaluation Metrics}
For evaluation, we adopt distortion-based and perceptual-based metrics. The distortion-based metrics include PSNR, SSIM, Y-PSNR, and Y-SSIM \cite{wang2004image}, while perceptual metrics include LPIPS \cite{zhang2018unreasonable} and FID \cite{heusel2017gans}. PSNR and SSIM evaluate the quality of the RGB channels in images. Y-PSNR and Y-SSIM assess the quality of the luminance (Y) channel in the YCbCr color space. For PSNR, SSIM, Y-PSNR, and Y-SSIM, higher values indicate the better performance, whereas for LPIPS and FID, lower values indicate better perceptual quality.

\subsubsection{Experimental Configurations}
To train the first stage of our model, we set the initial learning rate to 0.00003 and adjust it throughout the training process using a cosine annealing schedule. The model is trained in 300 epochs with a batch size of 256. A cosine similarity loss with a weight of 0.5 is employed to guide the training. The model is trained using four NVIDIA GeForce RTX 4090 GPUs. We select the CLIP ViT-B/32 model trained on LAION-2B, i.e., a 2-billion-sample English subset of LAION-5B \cite{schuhmann2022laion}, as the pretrained CLIP model. For training the second-stage image restoration model, the initial learning rate is set to 0.0001 and dynamically adjusted using a cosine annealing scheduler with $\beta_{1}=0.9$ and $\beta_{2}=0.99$. The model is trained for a total of 150,000 iterations with a batch size of 48. The patch size is 256 $\times$ 256. To enhance generalization, data augmentation techniques, such as random vertical and horizontal flips, are applied. Training is performed on six NVIDIA GeForce RTX 4090 GPUs. The Adam optimizer \cite{kingma2014adam} is used in both training stages.

\begin{figure*}
\centering
\includegraphics[width=\textwidth, height=23.5cm]{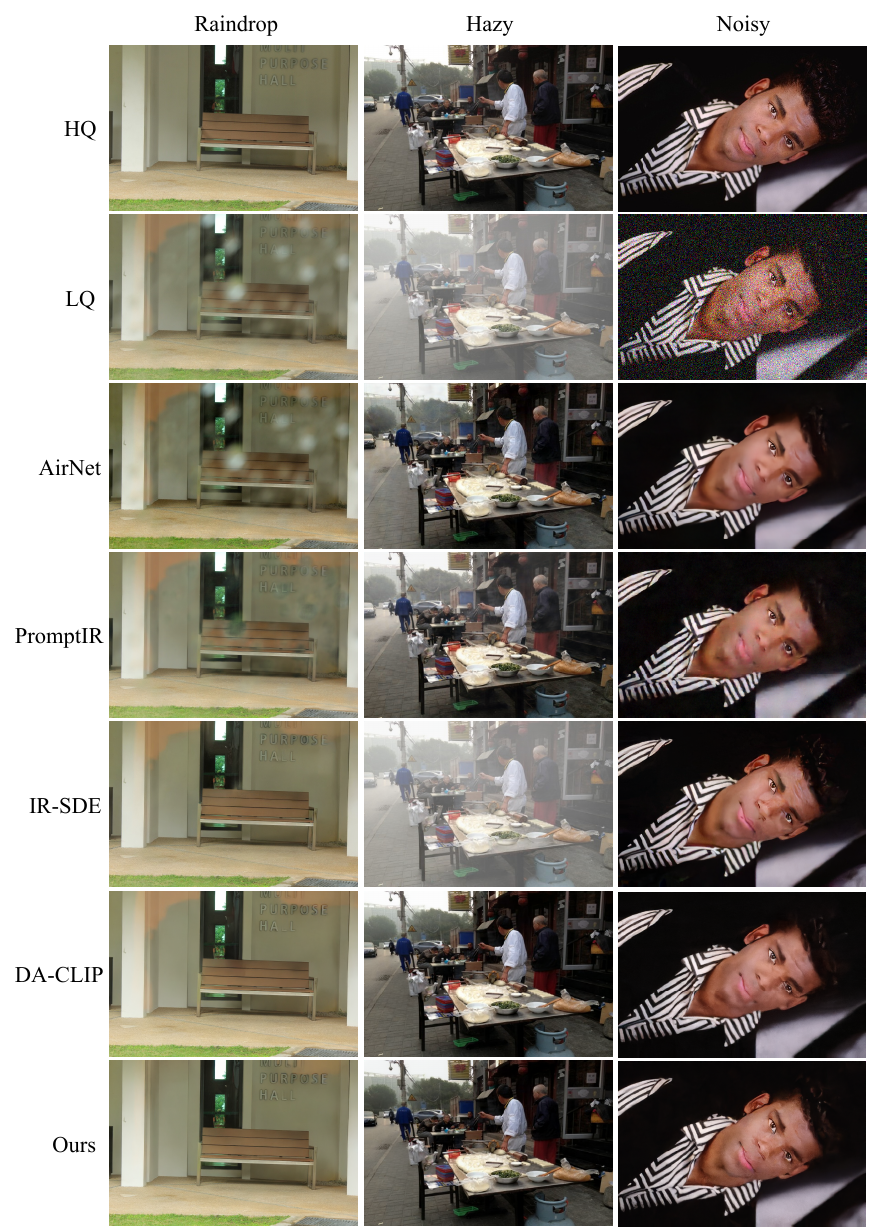}
\caption{Visual results of different universal image restoration methods across various image restoration tasks.}
\label{visual1}
\end{figure*}

\subsection{Comparisons with State-of-the-Art Methods}
\subsubsection{Universal Image Restoration}
We compare our proposed method with several state-of-the-art universal image restoration methods, including AirNet \cite{li2022all}, PromptIR \cite{potlapalli2023promptir}, IR-SDE \cite{luo2023image}, and DA-CLIP \cite{luo2023controlling}. Note that AirNet \cite{li2022all} is a convolutional neural network (CNN)-based model, while PromptIR \cite{potlapalli2023promptir} is built on Transformer architecture. Both AirNet and PromptIR use $l1$ loss for reconstruction, which guarantees pixel-level fidelity. IR-SDE \cite{luo2023image}, DA-CLIP \cite{luo2023controlling}, and our method are based on the generative diffusion-based models. 

Table~\ref{quan1} shows the quantitative performance of these methods across three image restoration tasks, i.e., raindrop removal, dehazing, and denoising. For raindrop removal and dehazing, our method achieves the best performance on all distortion-based metrics, i.e., PSNR, SSIM, Y-PSNR, and Y-SSIM, and competitive performance on LPIPS and FID. For denoising, although the CNN-based AirNet \cite{li2022all} and Transformer-based PromptIR \cite{potlapalli2023promptir} perform well on distortion-based metrics and LPIPS, their outputs tend to be overly smooth or visually unpleasing, as illustrated in Fig.~\ref{visual1}, leading to worse FID scores compared to generative model-based methods.

Fig.~\ref{visual1} shows the qualitative comparisons among universal image restoration methods on the raindrop removal, dehazing, and denoising tasks. AirNet \cite{li2022all} and PromptIR \cite{potlapalli2023promptir} struggle to remove raindrops. IR-SDE \cite{luo2023image} and DA-CLIP \cite{luo2023controlling} effectively eliminate most raindrops, but leave residual traces and struggle to address orange artifacts caused by raindrops on the camera lens. On the contrary, the proposed method removes all raindrops and the orange artifacts, restoring a clean image closest to the target HQ image. For dehazing, the proposed method can more effectively remove haze and generate more visually pleasing results. For denoising, as discussed above, AirNet and PromptIR generate overly smooth images due to the $l1$ loss constraint, achieving higher PSNR and SSIM. Our model, guided by visual and textual embeddings from CLIP and powered by a generative model, generates more realistic images with richer details.

\subsubsection{Degradation-specific Image Restoration}
To further evaluate the effectiveness of the proposed method, we assess our model not only in the universal image restoration setting, but also in a degradation-specific context. Specifically, we conduct experiments on the dehazing task. We compare the proposed method with state-of-the-art dehazing methods, including MAXIM \cite{tu2022maxim}, IR-SDE \cite{luo2023image}, and DA-CLIP \cite{luo2023controlling}. Unlike universal image restoration, degradation-specific methods are trained solely on the dehazing dataset. 

Table~\ref{quan2} shows a quantitative comparison of various dehazing methods. MAXIM \cite{tu2022maxim}, a transformer-based method, is trained using the Charbonnier loss \cite{barron2019general}, which is similar to $l1$ loss. Although MAXIM achieves high distortion-based evaluation metrics, it struggles to restore realistic and natural images due to the limitations of the $l1$ or $l2$ loss functions. The proposed method outperforms other generative model-based methods in PSNR and SSIM metrics, while also achieving superior perceptual quality in terms of LPIPS and FID. These results demonstrate the effectiveness and superiority of the proposed method.

\begin{figure*}
\centering
\includegraphics[width=\textwidth]{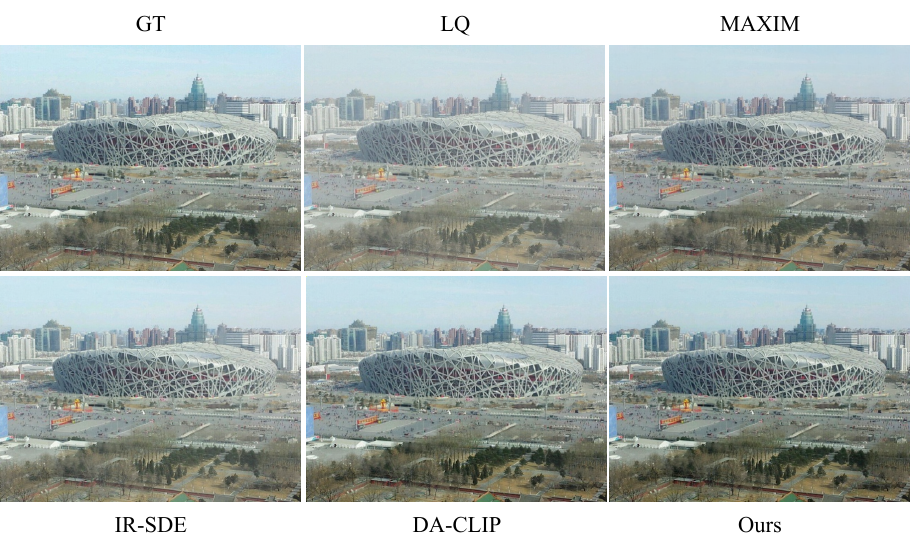}
\caption{Visual results of different dehazing methods on dehazing tasks.}
\label{visual2}
\end{figure*}

Fig.~\ref{visual2} presents the visual results of different dehazing methods. Compared to other methods, MAXIM \cite{tu2022maxim} eliminates the least amount of haze. IR-SDE \cite{luo2023image} also struggles to remove haze, especially around the trees. DA-CLIP \cite{luo2023controlling} generates clearer visual results than IR-SDE, though some haze remains on the buildings. Among all methods, the proposed method generates the clearest output with the richest details.

In summary, by combining visual perception and linguistic understanding, the proposed method can effectively learn from the guidance provided by the VLM to restore images affected by various isolated degradations.

\begin{table}
\centering
\caption{Quantitative comparison of different dehazing methods on the dehazing task. Best results are highlighted in red, and second-place results are marked in blue.}
\resizebox{.48\textwidth}{!}{\begin{tabular}{ccccccc}
\hline
Method  & PSNR                         & SSIM                          & Y-PSNR                       & Y-SSIM                        & LPIPS                         & FID                         \\ \hline\hline
MAXIM \cite{tu2022maxim}   & {\color[HTML]{FE0000} 31.05} & {\color[HTML]{FE0000} 0.9653} & {\color[HTML]{FE0000} 32.69} & {\color[HTML]{FE0000} 0.9814} & 0.0215                        & 9.34                        \\
IR-SDE \cite{luo2023image}  & 20.56                        & 0.8671                        & 22.03                        & 0.9152                        & 0.0718                        & 15.93                       \\
DA-CLIP \cite{luo2023controlling} & 30.18                        & 0.9513                        & 31.78                        & 0.9738                        & {\color[HTML]{3166FF} 0.0189} & {\color[HTML]{3166FF} 8.40} \\
Ours    & {\color[HTML]{3166FF} 30.34} & {\color[HTML]{3166FF} 0.9579} & {\color[HTML]{3166FF} 31.92} & {\color[HTML]{3166FF} 0.9757} & {\color[HTML]{FE0000} 0.0182} & {\color[HTML]{FE0000} 8.11} \\ \hline 
\end{tabular}}
\label{quan2}
\end{table}

\begin{table}
\centering
\caption{Comparison among various variants on the dehazing task. Best results are highlighted in bold.}
\resizebox{.48\textwidth}{!}{\begin{tabular}{ccccccc}
\hline
Variant  & PSNR           & SSIM            & Y-PSNR         & Y-SSIM          & LPIPS           & FID  \\ \hline\hline
SCA      & 29.51          & 0.9497          & 31.09          & 0.9741          & 0.0207          & 9.11  \\
ICA      & 29.67          & 0.9499          & 31.24          & 0.9738          & 0.0206          & \textbf{8.58}   \\
SCA\&ICA & \textbf{30.22} & \textbf{0.9523} & \textbf{31.85} & \textbf{0.9745} & \textbf{0.0199} & 8.99 \\ \hline
\end{tabular}}
\label{variant}
\end{table}

\begin{table}
\centering
\caption{Comparison of different text descriptions as input to the text encoder. `Null' indicates that no text is provided to the text encoder. Best results are highlighted in bold.}
\begin{tabular}{ccccc}
\hline
Text                & PSNR           & SSIM           & LPIPS           & FID            \\ \hline\hline
`This is a photo' & 29.90          & 0.9541          & \textbf{0.0190} & \textbf{8.572} \\ \hline
Null                & 29.55          & \textbf{0.9554} & 0.0198          & 8.86           \\ \hline
LQ caption          & \textbf{30.22} & 0.9513          & 0.0199          & 8.99           \\ \hline
\end{tabular}
\label{text}
\end{table}

\subsection{Ablation Studies}
\subsubsection{Effects of Modules}
The SCA and ICA modules are essential for processing textual and visual embeddings. 
To evaluate their impacts, we conduct ablation experiments on the dehazing task. Table~\ref{variant} shows the quantitative comparison of different model variants. `SCA' refers to the variant that includes only the SCA module, `ICA' includes only the ICA module, and `SCA\&ICA' contains both the SCA and ICA modules, i.e., our proposed model. The `ICA' variant performs better than `SCA' across all distortion-based and perceptual metrics, suggesting that image content plays a more important role than textual descriptions in image restoration. 
However, the full model with both SCA and ICA, outperforms all other variants on most evaluation metrics and achieves a comparable FID score. This highlights the significant contribution of both visual and textual embeddings to generating realistic and detailed restored images.

\subsubsection{Effects of Different LQ Descriptions}
We also study the impacts of different textual inputs to the text encoder. The LQ caption refers to the textual description of the LQ image content. As substitutes, we use `This is a photo' and a Null input (i.e., no text). Table~\ref{text} shows that omitting textual description input to the text encoder leads to a significant drop in pixel-level fidelity. Although the generic description `This is a photo' enhances the fidelity of the image compared to the Null input, it remains inferior to using actual LQ image captions. Perceptual metrics, such as LPIPS and FID, show slight degradation when using LQ captions, likely due to noise introduced by inaccurate descriptions of LQ images. Nevertheless, LQ captions enable the model to generate images with higher fidelity while achieving comparable perceptual quality, underscoring their importance in the image restoration process.

\section{Conclusion}
In this paper, we introduce the Vision-Language Model Guided Image Restoration (VLMIR) framework, which leverages the complementary visual and linguistic priors of vision-language models (VLMs) to advance universal image restoration. VLMIR employs a two-stage methodology: VLM-based feature extraction followed by diffusion-based image restoration. In the first stage, we align low-quality (LQ) and high-quality (HQ) caption embeddings using cosine similarity loss with LoRA fine-tuning, and apply contrastive learning to extract clean image content and degradation-specific embeddings. These features are then seamlessly integrated into a diffusion-based model (IR-SDE) via cross-attention mechanisms. Extensive experiments and ablation studies demonstrate that VLMIR outperforms existing methods in both universal and degradation-specific restoration tasks. Our findings underscore the critical role of integrating visual and linguistic knowledge from VLMs, paving the way for more generalizable and effective image restoration solutions.

\printcredits

\bibliographystyle{cas-model2-names}

\bibliography{cas-refs}



\end{document}